\documentclass[10pt,twocolumn,letterpaper]{article}

\usepackage[pagenumbers]{cvpr} 

\usepackage{graphicx}
\usepackage{amsmath}
\usepackage{amssymb}
\usepackage{booktabs}
\usepackage{romannum} 
\usepackage{algorithm} 
\usepackage{algpseudocode} 
\usepackage{wrapfig}
\usepackage{arydshln} 
\usepackage{makecell}
\usepackage{multirow}
\def\eg{\emph{e.g}\onedot{\emph{,~}}}  
\def\ie{\emph{i.e}\onedot{\emph{,~}}}  

\usepackage[pagebackref,breaklinks,colorlinks]{hyperref}

\usepackage[capitalize]{cleveref}
\crefname{section}{Sec.}{Secs.}
\Crefname{section}{Section}{Sections}
\Crefname{table}{Table}{Tables}
\crefname{table}{Tab.}{Tabs.}

\usepackage{amsmath,amsfonts,bm}

\def\eqref#1{equation~\ref{#1}}

\def\1{\bm{1}}

\DeclareMathAlphabet{\mathsfit}{\encodingdefault}{\sfdefault}{m}{sl}
\SetMathAlphabet{\mathsfit}{bold}{\encodingdefault}{\sfdefault}{bx}{n}

\newcommand{\E}{\mathbb{E}}

\DeclareMathOperator*{\argmin}{arg\,min}

\usepackage{dsfont}

\def\cvprPaperID{12113} 
\def\confName{CVPR}
\def\confYear{2023}

\begin{document}
\pagenumbering{arabic} 
\title{Demystifying Causal Features on Adversarial Examples and Causal Inoculation for Robust Network by Adversarial Instrumental Variable Regression}

\author{Junho Kim\thanks{Equal contribution. $\dagger$ Corresponding author.},~~Byung-Kwan Lee\footnote[1]{},~~Yong Man Ro\footnote[2]{}\\
Image and Video Systems Lab, School of Electrical Engineering, KAIST, South Korea\\
{\tt\small \{arkimjh, leebk, ymro\}@kaist.ac.kr}
}

\maketitle

\begin{abstract}
    The origin of adversarial examples is still inexplicable in research fields, and it arouses arguments from various viewpoints, albeit comprehensive investigations. In this paper, we propose a way of delving into the unexpected vulnerability in adversarially trained networks from a causal perspective, namely adversarial instrumental variable (IV) regression. By deploying it, we estimate the causal relation of adversarial prediction under an unbiased environment dissociated from unknown confounders. Our approach aims to demystify inherent causal features on adversarial examples by leveraging a zero-sum optimization game between a casual feature estimator (i.e., hypothesis model) and worst-case counterfactuals (i.e., test function) disturbing to find causal features. Through extensive analyses, we demonstrate that the estimated causal features are highly related to the correct prediction for adversarial robustness, and the counterfactuals exhibit extreme features significantly deviating from the correct prediction. In addition, we present how to effectively inoculate CAusal FEatures (CAFE) into defense networks for improving adversarial robustness.

\end{abstract}

\section{Introduction}
\label{sec:intro}
Adversarial examples, which are indistinguishable to human observers but maliciously fooling Deep Neural Networks (DNNs), have drawn great attention in research fields due to their security threats used to compromise machine learning systems. In real-world environments, such potential risks evoke weak reliability of the decision-making process for DNNs and pose a question of adopting DNNs in safety-critical areas~\cite{8756865, WANG201912, 8824956}.

To understand the origin of adversarial examples, seminal works have widely investigated the adversarial vulnerability through numerous viewpoints such as excessive linearity in a hyperplane~\cite{43405}, aberration of statistical fluctuations~\cite{42503, shafahi2018are}, and phenomenon induced from frequency information~\cite{yin2019fourier}. Recently, several works~\cite{NEURIPS2019_e2c420d9, kim2021distilling} have revealed the existence and pervasiveness of robust and non-robust features in adversarially trained networks and pointed out that the non-robust features on adversarial examples can provoke unexpected misclassifications.

\begin{figure}[t!]
\centering
\includegraphics[width=1.0\linewidth]{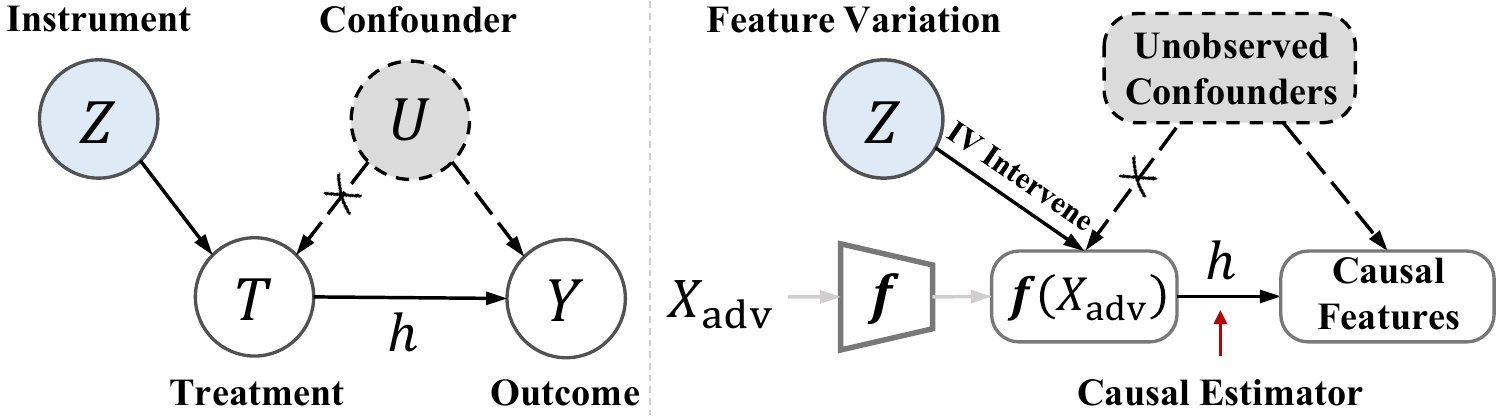}
\vspace*{-0.7cm}
\caption{Data generating process (DGP) with IV. By deploying $Z$, it can estimate causal relation between treatment $T$ and outcome $Y$ under exogenous condition for unknown confounders $U$.}
\label{fig:1}
\vspace{-0.6cm}
\end{figure}

Nonetheless, there still exists a lack of common consensus~\cite{unclear} on underlying causes of adversarial examples, albeit comprehensive endeavors~\cite{tsipras2018robustness, benchmarking}. It is because that the earlier works have focused on analyzing associations between adversarial examples and target labels in the learning scheme of adversarial training~\cite{madry2018towards, pmlr-v97-zhang19p, Wang2020Improving, wu2020adversarial, rade2022reducing}, which is canonical supervised learning. Such analyses easily induce spurious correlation (\ie statistical bias) in the learned associations, thereby cannot interpret the genuine origin of adversarial vulnerability under the existence of possibly biased viewpoints (\eg excessive linearity, statistical fluctuations, frequency information, and non-robust features). In order to explicate where the adversarial vulnerability comes from in a causal perspective and deduce true adversarial causality, we need to employ an intervention-oriented approach (\ie causal inference) that brings in estimating causal relations beyond analyzing merely associations for the given data population of adversarial examples.

One of the efficient tools for causal inference is instrumental variable (IV) regression when randomized controlled trials (A/B experiments) or full controls of unknown confounders are not feasible options. It is a popular approach used to identify causality in econometrics~\cite{newey2003instrumental, darolles2011nonparametric, chen2012estimation}, and it provides an unbiased environment from unknown confounders that raise the endogeneity of causal inference~\cite{reiersol1945confluence}. In IV regression, the instrument is utilized to eliminate a backdoor path derived from unknown confounders by separating exogenous portions of treatments. For better understanding, we can instantiate a case of finding causal relations~\cite{card1993using} between education $T$ and earnings $Y$ as illustrated in~\cref{fig:1}. Solely measuring correlation between the two variables does not imply causation, since there may exist unknown confounders $U$ (\eg individual ability, family background, etc.). Ideally, conditioning on $U$ is the best way to identify causal relation, but it is impossible to control the unobserved variables. David Card~\cite{card1993using} has considered IV as the college proximity $Z$, which is directly linked with education $T$ but intuitively not related with earnings $Y$. By assigning exogenous portion to $Z$, it can provide an unbiased environment dissociated from $U$ for identifying true causal relation between $T$ and $Y$.

Specifically, once regarding data generating process (DGP)~\cite{phillips1990statistical} for causal inference as in~\cref{fig:1}, the existence of unknown confounders $U$ could create spurious correlation generating a backdoor path that hinders causal estimator $h$ (\ie hypothesis model) from estimating causality between treatment $T$ and outcome $Y$ ($T$ $\leftarrow$ $U$ $\rightarrow$ $Y$). By adopting an instrument $Z$, we can acquire the estimand of true causality from $h$ in an unbiased state ($Z$ $\rightarrow$ $T$ $\rightarrow$ $Y$). Bringing such DGP into adversarial settings, the aforementioned controversial perspectives (\eg excessive linearity, statistical fluctuations, frequency information, and non-robust features) can be regarded as possible candidates of unknown confounders $U$ to reveal adversarial origins. In most observational studies, everything is endogenous in practice so that we cannot explicitly specify all confounders and conduct full controls of them in adversarial settings. Accordingly, we introduce IV regression as a powerful causal approach to uncover adversarial origins, due to its capability of causal inference although unknown confounders remain. 

Here, unknown confounders $U$ in adversarial settings easily induce ambiguous interpretation for the adversarial origin producing spurious correlation between adversarial examples and their target labels. In order to uncover the adversarial causality, we first need to intervene on the intermediate feature representation derived from a network $f$ and focus on what truly affects adversarial robustness irrespective of unknown confounders $U$, instead of model prediction. To do that, we define the instrument $Z$ as feature variation in the feature space of DNNs between adversarial examples and natural examples, where the variation $Z$ is originated from the adversarial perturbation in the image domain such that $Z$ derives adversarial features $T$ for the given natural features. Note that regarding $Z$ as instrument is reasonable choice, since the feature variation alone does not serve as relevant information for adversarial prediction without natural features. Next, once we find causality-related feature representations on adversarial examples, then we name them as \textit{causal features} $Y$ that can encourage robustness of predicting target labels despite the existence of adversarial perturbation as in ~\cref{fig:1}.

In this paper, we propose \textit{adversarial instrumental variable (IV) regression} to identify causal features on adversarial examples concerning the causal relation of adversarial prediction. Our approach builds an unbiased environment for unknown confounders $U$ in adversarial settings and estimates inherent causal features on adversarial examples by employing generalized method of moments (GMM)~\cite{hansen1982large} which is a flexible estimation for non-parametric IV regression. Similar to the nature of adversarial learning~\cite{gan, wgan}, we deploy a zero-sum optimization game~\cite{lewis2018adversarial, dikkala2020minimax} between a hypothesis model and test function, where the former tries to unveil causal relation between treatment and outcome, while the latter disturbs the hypothesis model from estimating the relation. In adversarial settings, we regard the hypothesis model as a causal feature estimator which extracts causal features in the adversarial features to be highly related to the correct prediction for the adversarial robustness, while the test function makes worst-case counterfactuals (\ie extreme features) compelling the estimand of causal features to significantly deviate from correct prediction. Consequently, it can further strengthen the hypothesis model to demystify causal features on adversarial examples.

Through extensive analyses, we corroborate that the estimated causal features on adversarial examples are highly related to correct prediction for adversarial robustness, and the test function represents the worst-case counterfactuals on adversarial examples. By utilizing feature visualization~\cite{mahendran2015understanding, olah2017feature}, we interpret the causal features on adversarial examples in a human-recognizable way. Furthermore, we introduce an inversion of the estimated causal features to handle them on the possible feature bound and present a way of efficiently injecting these \textit{CAusal FEatures (CAFE)} into defense networks for improving adversarial robustness.

\section{Related Work}
\label{sec:related}

In the long history of causal inference, there have been a variety of works~\cite{garcia2006causal, kim2009causal, hagmayer2017causal} to discover how the causal knowledge affects decision-making process. Among various causal approaches, especially in economics, IV regression~\cite{reiersol1945confluence} provides a way of identifying the causal relation between the treatment and outcome of interests despite the existence of unknown confounders, where IV makes the exogenous condition of treatments thus provides an unbiased environment for the causal inference.

Earlier works of IV regression~\cite{angrist1996identification, angrist2008mostly} have limited the relation for causal variables by formalizing it with linear function, which is known as 2SLS estimator~\cite{wooldridge2010econometric}. With progressive developments of machine learning methods, researchers and data scientists desire to deploy them for non-parametric learning~\cite{newey2003instrumental, darolles2011nonparametric, chen2012estimation, chen2018optimal} and want to overcome the linear constraints in the functional relation among the variables. As extensions of 2SLS, DeepIV~\cite{hartford2017deep}, KernelIV~\cite{singh2019kernel}, and Dual IV~\cite{muandet2020dual} have combined DNNs as non-parametric estimator and proposed effective ways of exploiting them to perform IV regression. More recently, generalized method of moments (GMM)~\cite{lewis2018adversarial, bennett2019deep, dikkala2020minimax} has been cleverly proposed a solution for dealing with the non-parametric hypothesis model on the high-dimensional treatments through a zero-sum optimization, thereby successfully achieving the non-parametric IV regression.

In parallel with the various causal approaches utilizing IV, uncovering the origin of adversarial examples is one of the open research problems that arouse controversial issues. In the beginning, \cite{43405} have argued that the excessive linearity in the networks' hyperplane can induce adversarial vulnerability. Several works~\cite{42503, shafahi2018are} have theoretically analyzed such origin as a consequence of statistical fluctuation of data population, or the behavior of frequency information in the inputs~\cite{yin2019fourier}. Recently, the existence of non-robust features in DNNs~\cite{NEURIPS2019_e2c420d9, kim2021distilling} is contemplated as a major cause of adversarial examples, but it still remains inexplicable~\cite{unclear}. 

Motivated by IV regression, we propose a way of estimating inherent causal features in adversarial features easily provoking the vulnerability of DNNs. To do that, we deploy the zero-sum optimization based on GMM between a hypothesis model and test function~\cite{lewis2018adversarial, bennett2019deep, dikkala2020minimax}. Here, we assign the role of causal feature estimator to hypothesis model and that of generating worst-case counterfactuals to test function disturbing to find causal features. This strategy results in learning causal features to overcome all trials and tribulations regarded as various types of adversarial perturbation. 


\section{Adversarial IV Regression}
\label{sec:proposed}

Our major goal is estimating inherent causal features on adversarial examples highly related to the correct prediction for adversarial robustness by deploying IV regression. Before identifying causal features, we first specify problem setup of IV regression and revisit non-parametric IV regression with generalized method of moments (GMM).

\noindent\textbf{Problem Setup.~}  We start from conditional moment restriction (CMR)~\cite{chamberlain1987asymptotic, ai2003efficient} bringing in an asymptotically efficient estimation with IV, which reduces spurious correlation (\ie statistical bias) between treatment $T$ and outcome of interest $Y$ caused by unknown confounders $U$~\cite{pearl2009causality} (see their relationships in~\cref{fig:1}). Here, the formulation of CMR can be written with a hypothesis model $h$, so-called a causal estimator on the hypothesis space $\mathcal{H}$ as follows:
\begin{equation}
\label{eq:cmr}
    \E_{T}[\psi_{T}(h)\mid Z]=\textbf{0},
\end{equation}
where $\psi_{T}:\mathcal{H}\rightarrow \mathbb{R}^{d}$ denotes a generalized residual function~\cite{chen2012estimation} on treatment $T$, such that it represents $\psi_{T}(h)=Y-h(T)$ considered as an outcome error for regression task. Note that $\textbf{0}\in\mathbb{R}^{d}$ describes zero vector and $d$ indicates the dimension for the outcome of interest $Y$, and it is also equal to that for the output vector of the hypothesis model $h$. The treatment is controlled for being exogenous~\cite{nizalova2016exogenous} by the instrument. In addition, for the given instrument $Z$, minimizing the magnitude of the generalized residual function $\psi$ implies asymptotically restricting the hypothesis model $h$ not to deviate from $Y$, thereby eliminating the internal spurious correlation on $h$ from the backdoor path induced by confounders $U$.

\subsection{Revisiting Non-parametric IV regression}

Once we find a hypothesis model $h$ satisfying CMR with instrument $Z$, we can perform IV regression to endeavor causal inference using $h$ under the following formulation: $\E_{T}[h(T)\mid Z]=\int_{t \in T}h(t)\mathrm{d}\mathbb{P}(T=t\mid Z)$, where $\mathbb{P}$ indicates a conditional density measure. In fact, two-stage least squares (2SLS)~\cite{angrist1996identification, angrist2008mostly, wooldridge2010econometric} is a well-known solver to expand IV regression, but it cannot be directly applied to more complex model such as non-linear model, since 2SLS is designed to work on linear hypothesis model~\cite{peters2017elements}. Later, \cite{hartford2017deep} and \cite{singh2019kernel} 
have introduced a generalized 2SLS for non-linear model by using a conditional mean embedding and a mixture of Gaussian, respectively. Nonetheless, they still raise an ill-posed problem yielding biased estimates~\cite{bennett2019deep, muandet2020dual, dikkala2020minimax, zhang2020maximum} with the non-parametric hypothesis model $h$ on the high dimensional treatment $T$, such as DNNs. It stems from the curse nature of two-stage methods, known as \textit{forbidden regression}~\cite{angrist2008mostly} according to Vapnik’s principle~\cite{de2018statistical}: \textit{``do not solve
a more general problem as an intermediate step''}.

To address it, recent studies~\cite{lewis2018adversarial, bennett2019deep, dikkala2020minimax} have employed generalized method of moments (GMM) to develop IV regression and achieved successful one-stage regression alleviating biased estimates. Once we choose a moment to represent a generic outcome error with respect to the hypothesis model and its counterfactuals, GMM uses the moment to deliver infinite moment restrictions to the hypothesis model, beyond the simple constraint of CMR. Expanding \cref{eq:cmr}, the formulation of GMM can be written with a moment, denoted by $m:\mathcal{H}\times\mathcal{G}\rightarrow \mathbb{R}$ as follows (see Appendix A):
\begin{equation}
\label{eq:moment}
    \begin{aligned}
        m(h, g)&=\E_{Z, T}[\psi_{T}(h)\cdot g(Z)]
        \\&=\E_{Z}[\underbrace{\E_{T}[\psi_{T}(h)\mid Z]}_{\text{CMR}}\cdot g(Z)]=0,
        \vspace{-1mm}
    \end{aligned}
\end{equation}
where the operator $\cdot$ specifies inner product, and $g\in \mathcal{G}$ denotes test function that plays a role in generating infinite moment restrictions on test function space $\mathcal{G}$, such that its output has the dimension of $\mathbb{R}^{d}$. The infinite number of test functions expressed by arbitrary vector-valued functions $\{g_{1}, g_{2}, \cdots\} \in \mathcal{G}$ cues potential moment restrictions (\ie empirical counterfactuals)~\cite{blundell2001estimation} violating~\cref{eq:moment}. In other words, they make it easy to capture the worst part of IV which easily stimulates the biased estimates for hypothesis model $h$, thereby helping to obtain more genuine causal relation from $h$ by considering all of the possible counterfactual cases $g$ for generalization.

However, it has an analogue limitation that we cannot deal with infinite moments because we only handle observable finite number of test functions. Hence, recent studies construct maximum moment restriction~\cite{dikkala2020minimax, zhang2020maximum, muandet2020kernel} to efficiently tackle the infinite moments by focusing only on the extreme part of IV, denoted as $\sup_{g \in \mathcal{G}}m(h,g)$ in a closed-form expression. By doing so, we can concurrently minimize the moments for the hypothesis model to fully satisfy the worst-case generalization performance over test functions. Thereby, GMM can be re-written with min-max optimization thought of as a zero-sum game between the hypothesis model $h$ and test function $g$:
\begin{equation}
\label{eq:gmm}   \min\limits_{h\in\mathcal{H}}\sup\limits_{g\in\mathcal{G}}m(h,g)\approx\min\limits_{h\in\mathcal{H}}\max\limits_{g\in\mathcal{G}}\E_{Z, T}[\psi_{T}(h) \cdot g(Z)],
\end{equation}
where the infinite number of test functions can be replaced with the non-parametric test function in the form of DNNs. Next, we bridge GMM of~\cref{eq:gmm} to adversarial settings and unveil the adversarial origin by establishing adversarial IV regression with maximum moment restriction.

\subsection{Demystifying Adversarial Causal Features}

To demystify inherent causal features on adversarial examples, we first define feature variation $Z$ as the instrument, which can be written with adversarially trained DNNs denoted by $f$ as follows:
\begin{equation}
\label{eq:inst}
    Z=f_{l}(X_\epsilon)-f_{l}(X)=F_{\text{adv}}-F_{\text{natural}},
\end{equation}
where $f_{l}$ outputs a feature representation in $l^{\text{th}}$ intermediate layer, $X$ represents natural inputs, and $X_\epsilon$ indicates adversarial examples with adversarial perturbation $\epsilon$ such that $X_{\epsilon}=X+\epsilon$. In the sense that we have a desire to uncover how adversarial features $F_{\text{adv}}$ truly estimate causal features $Y$ which are outcomes of our interests, we set the treatment to $T=F_{\text{adv}}$ and set counterfactual treatment with a test function to $T_{\text{CF}}=F_{\text{natural}}+g(Z)$.

Note that, if we na\"{\i}vely apply test function $g$ to adversarial features $T$ to make counterfactual treatment $T_{\text{CF}}$ such that $T_{\text{CF}}=g(T)$, then the outputs (\ie causal features) of hypothesis model $h(T_{\text{CF}})$ may not be possibly acquired features considering feature bound of DNNs $f$. In other words, if we do not keep natural features in estimating causal features, then the estimated causal features will be too exclusive features from natural ones. This results in non-applicable features considered as an imaginary feature we cannot handle, since the estimated causal features are significantly manipulated ones only in a specific intermediate layer of DNNs. Thus, we set counterfactual treatment to $T_{\text{CF}}=F_{\text{natural}}+g(Z)$. This is because above formation can preserve natural features, where we first subtract natural features from counterfactual treatment such that $T'=T_{\text{CF}}-F_{\text{natural}}=g(Z)$ and add the output $Y'$ of hypothesis model to natural features for recovering causal features such that $Y=Y'+F_{\text{natural}}=h(T')+F_{\text{natural}}$. In brief, we intentionally translate causal features and counterfactual treatment not to deviate from possible feature bound.

Now, we newly define \textit{Adversarial Moment Restriction (AMR)} including the counterfactuals computed by the test function for adversarial examples, as follows: $\E_{T'}[\psi_{T'}(h)\mid Z]=\textbf{0}$. Here, the generalized residual function $\psi_{T'\mid Z}(h)=Y'-h(T')$ in adversarial settings deploys the translated causal features $Y'$. Bring them together, we re-formulate GMM with counterfactual treatment to fit adversarial IV regression, which can be written as (Note that $h$ and $g$ consist of a simple CNN structure):
\begin{equation}
\label{eq:amr gmm}
    \min\limits_{h\in\mathcal{H}}\max\limits_{g\in\mathcal{G}}\E_{Z}[\underbrace{\E_{T'}[\psi_{T'}(h)\mid Z]}_{\text{AMR}}g(Z)]=\E_{Z}[\psi_{T'\mid Z}(h)g(Z)],
\end{equation}
where it satisfies $\E_{T'}[\psi_{T'}(h)\mid Z]=\psi_{T'\mid Z}(h)$ because $Z$ corresponds to only one translated counterfactual treatment $T'=g(Z)$. Here, we cannot directly compute the generalized residual function $\psi_{T'\mid Z}(h)=Y'-h(T')$ in AMR, since there are no observable labels for the translated causal features $Y'$ on high-dimensional feature space. Instead, we make use of onehot vector-valued target label $G\in\mathbb{R}^{K}$ ($K$ : class number) corresponding to the natural input $X$ in classification task. To utilize it, we alter the domain of computing GMM from feature space to log-likelihood space of model prediction by using the log-likelihood function: $\Omega(\omega)=\log f_{l+}(F_{\text{natural}}+\omega)$, where $f_{l+}$ describes the subsequent network returning classification probability after $l^{\text{th}}$ intermediate layer. Accordingly, the meaning of our causal inference is further refined to find inherent causal features of correctly predicting target labels even under worst-case counterfactuals. To realize it,~\cref{eq:amr gmm} is modified with moments projected to the log-likelihood space  as follows:
\begin{equation}
\label{eq:log amr gmm}
    \begin{aligned}
        \min\limits_{h\in\mathcal{H}}\max\limits_{g\in\mathcal{G}}&\E_{Z}[\psi^{\Omega}_{T'\mid Z}(h)\cdot (\Omega\circ g)(Z)]\\&=\E_{Z}[\{G_{\log}-(\Omega\circ h)(T')\} \cdot (\Omega\circ g)(Z)],
    \end{aligned}
\end{equation}
where $\psi^{\Omega}_{T'\mid Z}(h)$ indicates the generalized residual function on the log-likelihood space, the operator $\circ$ symbolizes function composition, and $G_{\log}$ is log-target label such that satisfies $G_{\log}=\log G$. Each element ($k=1,2,\cdots,K$) of log-target label has $G_{\log}^{(k)}=0$ when it is $G^{(k)}=1$ and has $G_{\log}^{(k)}=-\infty$ when it is $G^{(k)}=0$. To implement it, we just ignore the element $G_{\log}^{(k)}=-\infty$ and use another only.

So far, we construct GMM based on AMR in~\cref{eq:log amr gmm}, namely \textit{AMR-GMM}, to behave adversarial IV regression. However, there is absence of explicitly regularizing the test function, thus there happens generalization gap between ideal and empirical moments (see Appendix B). Thereby, it violates possible feature bounds of the test function and brings in imbalanced predictions on causal inference (see \cref{fig:rad}). To become a rich test function, previous works~\cite{lewis2018adversarial, bennett2019deep, dikkala2020minimax, wang2021scalable} have employed \textit{Rademacher complexity}~\cite{bartlett2002rademacher,koltchinskii2002empirical,yin2019rademacher} that provides tight generalization bounds for a family of functions. It has a strong theoretical foundation to control a generalization gap, thus it is related to various regularizers used in DNNs such as weight decay, Lasso, Dropout, and Lipschitz~\cite{wan2013regularization, zhai2018adaptive, pmlr-v80-du18a, NEURIPS2019_0e795480}. In AMR-GMM, it plays a role in enabling the test functions to find out the worst-case counterfactuals within adversarial feature bound. Following Appendix B, we build a final objective of AMR-GMM with rich test function as follows:
\begin{equation}
    \label{eq:final amr gmm}
    \min\limits_{h\in\mathcal{H}}\max\limits_{g\in\mathcal{G}}\E_{Z}[\psi^{\Omega}_{T'\mid Z}(h)\cdot (\Omega\circ g)(Z)]-|\E_{Z}[Z-g(Z)]|^2.
\end{equation}
Please see more details of AMR-GMM algorithm attached in Appendix D due to page limits. 

\section{Analyzing Properties of Causal Features}

In this section, we first notate several conjunctions of feature representation from the result of adversarial IV regression with AMR-GMM as: (\lowercase\expandafter{\romannumeral1}) \textit{Adversarial Feature} (Adv): $F_{\text{natural}}+Z$, (\lowercase\expandafter{\romannumeral2}) \textit{CounterFactual Feature} (CF): $F_{\text{natural}}+g(Z)$, (\lowercase\expandafter{\romannumeral3}) \textit{Counterfactual Causal Feature} (CC): $F_{\text{natural}}+(h\circ g)(Z)$, and (\lowercase\expandafter{\romannumeral4}) \textit{Adversarial Causal Feature} (AC): $F_{\text{natural}}+h(Z)$. By using them, we estimate adversarial robustness computed by classification accuracy for which the above feature conjunctions are propagated through $f_{l+}$, where standard attacks generate feature variation $Z$ and adversarial features $T$. Note that, implementation of all feature representations is treated at the last convolutional layer of DNNs $f$ as in~\cite{kim2021distilling}, since it mostly contains the high-level object concepts and has the unexpected vulnerability for adversarial perturbation due to high-order interactions~\cite{deng2022discovering}. Here, average treatment effects (ATE)~\cite{holland1986statistics}, used for conventional validation of causal approach, is replaced with adversarial robustness of the conjunctions.

\subsection{Validating Hypothesis Model and Test Function}
After optimizing hypothesis model and test function using AMR-GMM for adversarial IV regression, we can then control endogenous treatment (\ie adversarial features) and separate exogenous portion from it, namely causal features, in adversarial settings. Here, the hypothesis model finds causal features on adversarial examples, highly related to correct prediction for adversarial robustness even with the adversarial perturbation. On the other hand, the test function generates worst-case counterfactuals to disturb estimating causal features, thereby degrading capability of hypothesis model. These learning strategy enables hypothesis model to estimate inherent causal features overcoming all trials and tribulations from the counterfactuals. Therefore, the findings of the causal features on adversarial examples has theoretical evidence by nature of AMR-GMM to overcome various types of adversarial perturbation. Note that, our IV setup posits homogeneity assumption~\cite{heckman2006understanding}, a more general version than monotonicity assumption~\cite{angrist1996identification}, that adversarial robustness (\ie average treatment effects) consistently retains high for all data samples despite varying natural features $F_{\text{natural}}$ depending on data samples.

\begin{figure}[t!]
\centering
\includegraphics[width=0.99\linewidth]{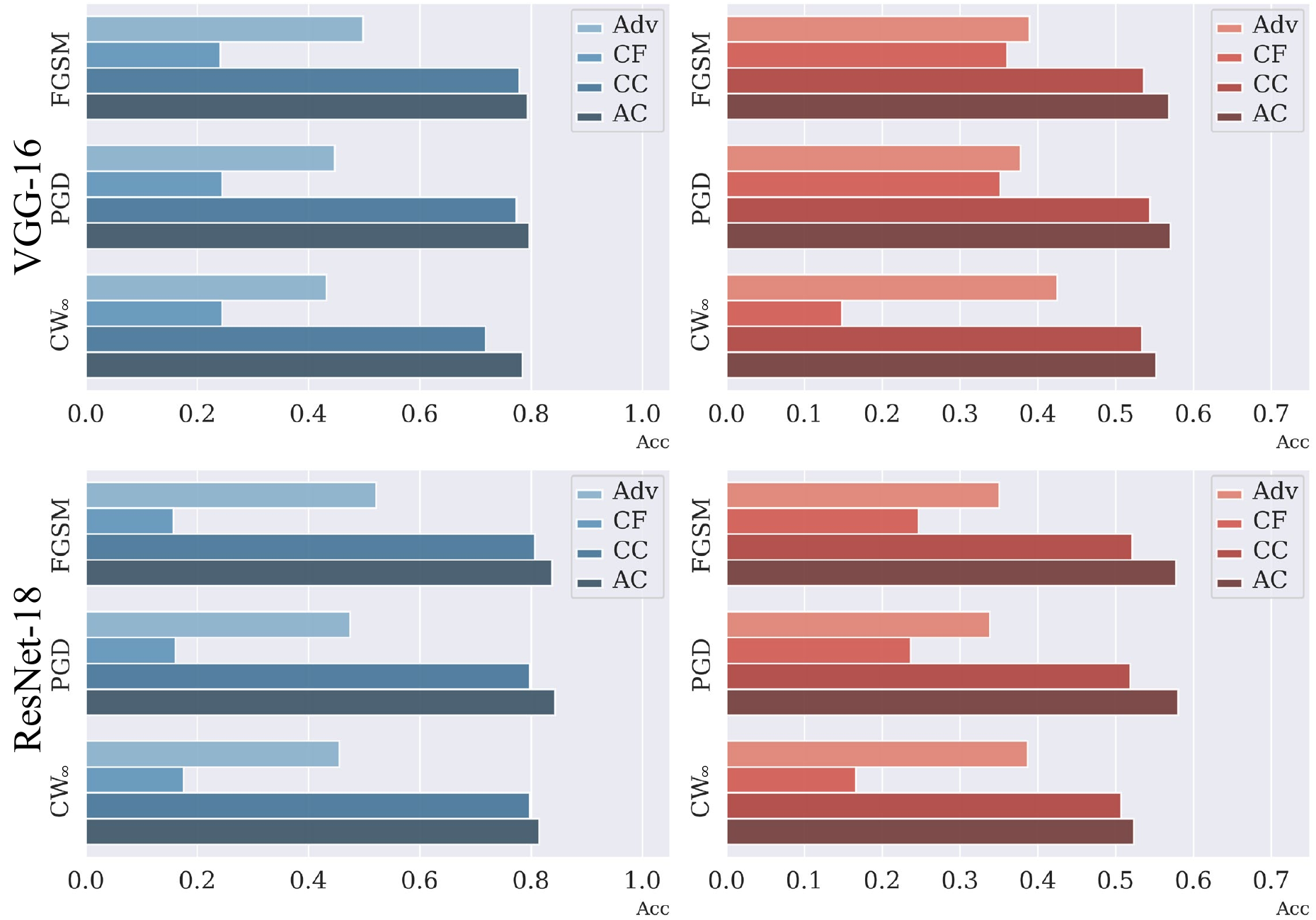}
\vspace*{-0.6cm}
\begin{flushleft}
    \hspace{1.3cm}{(a) CIFAR-10\hspace{2.1cm}(b) ImageNet}
\end{flushleft}
\vspace*{-0.4cm}
\caption{Adversarial robustness of Adv, CF, CC, AC on VGG-16 and ResNet-18 under three attack modes: FGSM\cite{43405}, PGD~\cite{madry2018towards}, CW$_{\infty}$~\cite{CW} for CIFAR-10~\cite{krizhevsky2009learning} and ImageNet~\cite{deng2009imagenet}.}
\label{fig:2}
\vspace{-0.6cm}
\end{figure}

As illustrated in~\cref{fig:2}, we intensively examine the average treatment effects (\ie adversarial robustness) for the hypothesis model and test function by measuring classification accuracy of the feature conjunctions (\ie Adv, CF, CC, AC) for all dataset samples. Here, we observe that the adversarial robustness of CF is inferior to that of CC, AC, and even Adv. Intuitively, it is an obvious result since the test function violating~\cref{eq:final amr gmm} forces feature representation to be the worst possible condition of extremely deviating from correct prediction. For the prediction results for CC and AC, they show impressive robustness performance than Adv with large margins. Since AC directly leverages the feature variation acquired from adversarial perturbation, they present better adversarial robustness than CC obtained from the test function outputting the worst-case counterfactuals on the feature variation. Intriguingly, we notice that both results from the hypothesis model generally show constant robustness even in a high-confidence adversarial attack~\cite{CW} fabricating unseen perturbation. Such robustness demonstrates the estimated causal features have ability to overcome various types of adversarial perturbation.

\begin{figure*}[t!]
\centering
\includegraphics[width=0.9\textwidth]{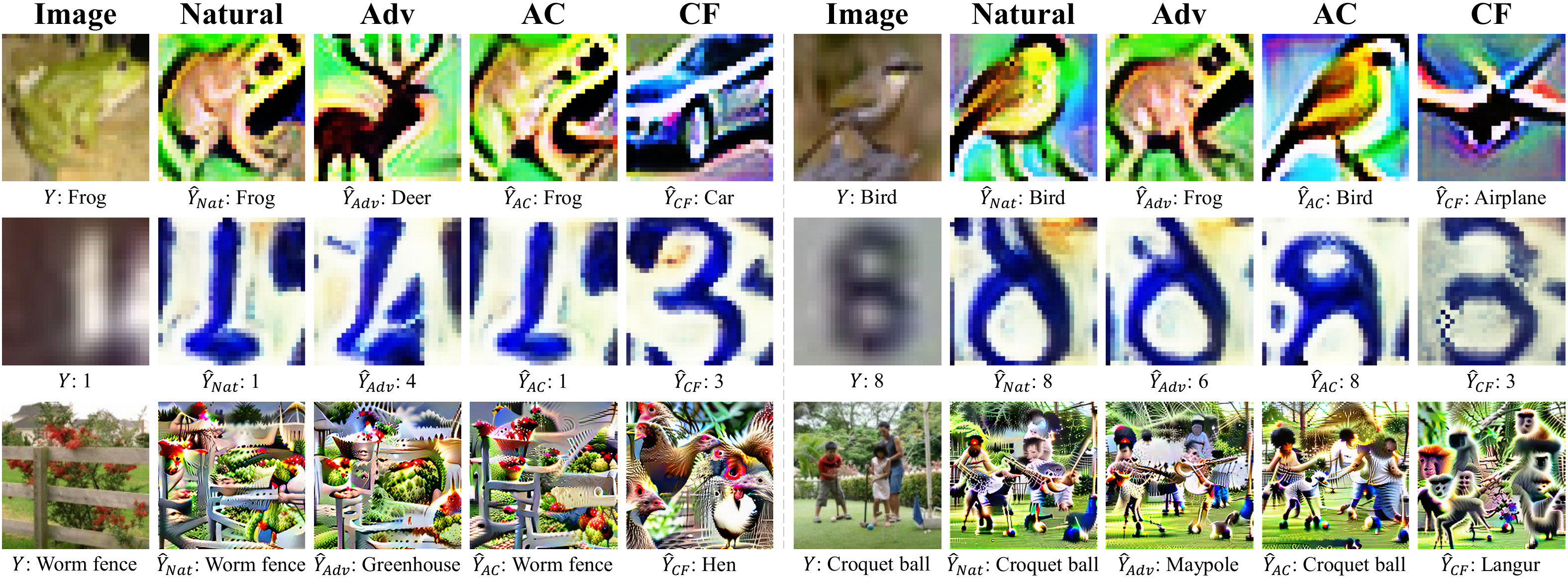}
\vspace*{-0.2cm}
\caption{Feature visualization results of representing natural features, Adv, AC, and CF. From the top row, CIFAR-10, SVHN, and ImageNet are sequentially used for the feature visual interpretation.}
\vspace*{-0.4cm}
\label{fig:3}
\end{figure*}

\subsection{Interpreting Causal Effects and Visual Results}
We have reached the causal features in adversarial examples and analyzed their robustness. After that, our next question is \textit{"Can the causal features per se have semantic information for target objects?"}. Recent works~\cite{engstrom2019adversarial, kim2021distilling, lee2022masking} have investigated to figure out the semantic meaning of feature representation in adversarial settings, we also utilize the feature visualization method~\cite{mahendran2015understanding, olah2017feature, nguyen2019understanding} on the input domain to interpret the feature conjunctions in a human-recognizable manner. As shown in~\cref{fig:3}, we can generally observe that the results of natural features represent semantic meaning of target objects. On the other hand, adversarial features (Adv) compel its feature representation to the orient of adversarially attacked target objects.

As aforementioned, the test function distracts treatments to be worst-case counterfactuals, which exacerbates the feature variation from adversarial perturbation. Thereby, the visualization of CF is remarkably shifted to the violated feature representation for target objects. For instance, as in ImageNet~\cite{deng2009imagenet} examples, we can see that the visualization of CF displays \textit{Hen} and \textit{Langur} features, manipulated from \textit{Worm fence} and \textit{Croquet ball}, respectively. We note that red flowers in original images have changed into red cockscomb and patterns of hen feather, in addition, people either have changed into distinct characteristics of langur, which accelerates the disorientation of feature representation to the worst counterfactuals. Contrastively, the visualization of AC displays a prominent exhibition and semantic consistency for target objects, where we can recognize their semantic information by themselves and explicable to human observers. By investigating visual interpretations, we reveal that feature representations acquired from the hypothesis model and test function both have causally semantic information, and their roles are in line with the theoretical evidence of our causal approach. In brief, we validate semantic meaning of causal features immanent in high-dimensional space despite the counterfactuals.

\subsection{Validating Conditions of IV Setup}

The instrumental variable needs to satisfy the following three valid conditions in order to successfully achieve non-parametric IV regression based on previous works \cite{hartford2017deep, muandet2020dual}: independent of the outcome error such that $\psi \perp Z$ (\textit{Unconfoundedness}) where $\psi$ denotes outcome error, and do not directly affect outcomes such that $Z \perp Y \mid T, \psi$ (\textit{Exclusion Restriction}) but only affect outcomes through a connection of treatments such that $\mathrm{Cov}(Z, T)\neq0$ (\textit{Relevance}).

For \textit{Unconfoundedness}, various works~\cite{madry2018towards, pmlr-v97-zhang19p, Wang2020Improving, wu2020adversarial, rade2022reducing} have proposed adversarial training robustifying DNNs $f$ with adversarial examples inducing feature variation that we consider as IV to improve robustness. In other words, when we see them in a perspective of IV regression, we can regard them as the efforts satisfying CMR in DNNs $f$ for the given feature variation $Z$. Aligned with our causal viewpoints, the first row in~\cref{tab:ivsetup} shows the existence of adversarial robustness with adversarial features $T$. Therefore, we can say that our IV (\ie feature variation) on adversarially trained models satisfies valid condition of \textit{Unconfoundedness}, so that IV is independent of the outcome error.

For \textit{Exclusion Restriction}, feature variation $Z$ itself cannot serve as enlightening information to model prediction without natural features, because only propagating the residual feature representation has no effect to model prediction by the learning nature of DNNs. Empirically, the second row in~\cref{tab:ivsetup} demonstrates that $Z$ cannot be helpful representation for prediction. Thereby, our IV is not encouraged to be correlated directly with the outcome, and it satisfies valid condition of \textit{Exclusion Restriction}.

\begin{table}[t!]
\centering
\renewcommand{\tabcolsep}{1.7mm}
\resizebox{0.95\linewidth}{!}{
\begin{tabular}{cccccccccc}
\Xhline{3\arrayrulewidth}
\multicolumn{1}{c}{} & \multicolumn{3}{c}{VGG}         & \multicolumn{3}{c}{ResNet}      & \multicolumn{3}{c}{WRN}         \\
\cmidrule(lr){2-4}\cmidrule(lr){5-7}\cmidrule(lr){8-10}
\multicolumn{1}{c}{} & CIFAR & SVHN & Tiny & CIFAR & SVHN & Tiny & CIFAR & SVHN & Tiny \\
\midrule
$f_{l+}(T)$                        & 44.8     & 52.1 & 21.5          & 46.5     & 55.4 & 24.2          & 48.7     & 56.7 & 25.5          \\
\cdashline{1-10}\noalign{\vskip 0.5ex}
$f_{l+}(Z)$                        & 0.0      & 0.0  & 0.0           & 0.0      & 0.0  & 0.0           & 0.0      & 0.0  & 0.0           \\
\cdashline{1-10}\noalign{\vskip 0.5ex}
$\rho$                        & 0.9     & 0.8 & 0.8          & 0.9     & 0.8 & 0.7          & 0.9     & 0.9 & 0.8        \\
\Xhline{3\arrayrulewidth}
\end{tabular}
}
\vspace*{-0.2cm}
\caption{Empirical validation for three conditions of our IV setup. $f_{l+}(T)$ and $f_{l+}(Z)$ indicates model performance (\%) of adversarial robustness by propagating adversarial features $T$ and feature variation $Z$ with subsequent network, respectively. The last row represents Pearson correlation: $\rho=\mathrm{Cov}(Z,T)/\sigma_Z\sigma_T$.}
\label{tab:ivsetup}
\vspace*{-0.5cm}
\end{table}
\begin{table*}[t!]
\centering
\renewcommand{\tabcolsep}{2.0mm}
\resizebox{0.95\linewidth}{!}{
\begin{tabular}{clccccccccccccccccccccc}
\Xhline{3\arrayrulewidth}
\multicolumn{1}{l}{}     & \multirow{2}{*}{Method} & \multicolumn{7}{c}{CIFAR-10}                                                                                  & \multicolumn{7}{c}{SVHN}                                                                                      & \multicolumn{7}{c}{Tiny-ImageNet}                                                                             \\
\cmidrule(lr){3-9}\cmidrule(lr){10-16}\cmidrule(lr){17-23}
\multicolumn{1}{l}{}     &                         & Natural       & FGSM          & PGD           & CW$_{\infty}$    & AP            & DLR           & AA            & Natural       & FGSM          & PGD           & CW$_{\infty}$    & AP            & DLR           & AA            & Natural       & FGSM          & PGD           & CW$_{\infty}$    & AP            & DLR           & AA            \\
\midrule
\multirow{10}{*}{\rotatebox[origin=c]{90}{VGG}}{\hskip 1ex}    & ADV                     & \textbf{78.5} & 49.8          & 44.8          & 42.6          & 43.2          & 42.9          & 40.7          & \textbf{91.9} & 64.8          & 52.1          & 48.9          & 48.0          & 48.5          & 45.2          & \textbf{53.2} & 25.3          & 21.5          & 21.0          & 20.2          & 20.8          & 19.6          \\
                         & ADV$_{\text{CAFE}}$   & 78.4          & \textbf{52.2} & \textbf{47.9} & \textbf{44.1} & \textbf{46.4} & \textbf{44.5} & \textbf{42.7} & 91.5          & \textbf{67.0} & \textbf{55.3} & \textbf{50.0} & \textbf{51.3} & \textbf{49.6} & \textbf{46.1} & 52.6          & \textbf{26.0} & \textbf{22.8} & \textbf{22.1} & \textbf{21.8} & \textbf{22.0} & \textbf{21.0} \\
\cmidrule{2-23}
                         & TRADES                  & \textbf{79.5} & 50.4          & 45.7          & 43.2          & 44.4          & 42.9          & 41.8          & \textbf{91.9} & 66.4          & 53.6          & 49.1          & 49.1          & 47.7          & 45.2          & \textbf{52.8} & 25.9          & 22.5          & 21.9          & 21.5          & 21.8          & 20.7          \\
                         & TRADES$_{\text{CAFE}}$& 77.0          & \textbf{51.6} & \textbf{47.9} & \textbf{44.0} & \textbf{47.0} & \textbf{43.9} & \textbf{42.7} & 90.3          & \textbf{67.8} & \textbf{56.1} & \textbf{50.0} & \textbf{53.6} & \textbf{49.1} & \textbf{47.5} & 52.1          & \textbf{26.5} & \textbf{23.6} & \textbf{22.6} & \textbf{22.5} & \textbf{22.6} & \textbf{21.6} \\
\cmidrule{2-23}
                         & MART                    & \textbf{79.7} & 52.4          & 47.2          & 43.4          & 45.5          & 43.8          & 42.0          & \textbf{92.6} & 66.6          & 54.2          & 47.9          & 49.6          & 47.1          & 44.4          & \textbf{53.1} & 25.0          & 21.5          & 21.2          & 20.4          & 21.0          & 19.9          \\
                         & MART$_{\text{CAFE}}$  & 78.3          & \textbf{54.2} & \textbf{49.7} & \textbf{43.9} & \textbf{48.1} & \textbf{44.5} & \textbf{42.7} & 91.3          & \textbf{67.6} & \textbf{57.3} & \textbf{49.5} & \textbf{54.2} & \textbf{48.3} & \textbf{46.4} & 53.0          & \textbf{25.6} & \textbf{22.3} & \textbf{21.6} & \textbf{21.3} & \textbf{21.5} & \textbf{20.5} \\
\cmidrule{2-23}
                         & AWP                     & \textbf{78.0} & 51.7          & 48.2          & 43.5          & 47.2          & 43.4          & 42.6          & 90.8          & 65.5          & 56.6          & 50.4          & 54.0          & 49.7          & 48.6          & 52.6          & 28.0          & 25.7          & 23.6          & 24.8          & 23.5          & 22.8          \\
                         & AWP$_{\text{CAFE}}$   & 77.4          & \textbf{54.8} & \textbf{51.4} & \textbf{44.2} & \textbf{50.2} & \textbf{44.9} & \textbf{43.5} & \textbf{91.9} & \textbf{67.9} & \textbf{58.6} & \textbf{51.2} & \textbf{55.9} & \textbf{51.1} & \textbf{49.7} & \textbf{52.9} & \textbf{28.8} & \textbf{26.4} & \textbf{24.2} & \textbf{25.6} & \textbf{24.1} & \textbf{23.4} \\
\cmidrule{2-23}
                         & HELP                    & \textbf{77.4} & 51.8          & 48.3          & 43.9          & 47.3          & 43.9          & 42.9          & 91.2          & 65.8          & 56.6          & 50.9          & 53.9          & 50.2          & 48.8          & \textbf{53.0} & 28.3          & 25.9          & 23.9          & 25.1          & 23.8          & 23.1          \\
                         & HELP$_{\text{CAFE}}$  & 75.6          & \textbf{54.4} & \textbf{51.4} & \textbf{44.6} & \textbf{50.4} & \textbf{44.8} & \textbf{43.7} & \textbf{91.5} & \textbf{67.3} & \textbf{58.5} & \textbf{51.6} & \textbf{56.2} & \textbf{51.4} & \textbf{50.0} & 52.6          & \textbf{29.4} & \textbf{27.1} & \textbf{24.7} & \textbf{26.4} & \textbf{24.4} & \textbf{23.9} \\
\midrule
\multirow{10}{*}{\rotatebox[origin=c]{90}{ResNet}}{\hskip 1ex} & ADV                     & 82.0          & 52.1          & 46.5          & 44.8          & 44.8          & 44.8          & 43.0          & \textbf{92.8} & 70.4          & 55.4          & 51.3          & 50.9          & 51.0          & 47.5          & \textbf{57.2} & 27.3          & 24.2          & 23.2          & 22.8          & 23.2          & 21.8          \\
                         & ADV$_{\text{CAFE}}$   & \textbf{82.6} & \textbf{55.9} & \textbf{50.7} & \textbf{47.6} & \textbf{49.0} & \textbf{47.7} & \textbf{46.2} & 92.5          & \textbf{73.6} & \textbf{58.9} & \textbf{53.8} & \textbf{54.9} & \textbf{52.6} & \textbf{49.8} & 56.3          & \textbf{28.6} & \textbf{25.7} & \textbf{24.7} & \textbf{24.4} & \textbf{24.6} & \textbf{23.5} \\
\cmidrule{2-23}
                         & TRADES                  & \textbf{83.0} & 55.0          & 49.8          & 47.5          & 48.3          & 47.3          & 46.1          & \textbf{93.2} & 72.8          & 57.7          & 52.6          & 53.0          & 51.5          & 48.9          & \textbf{56.5} & 28.4          & 25.3          & 24.4          & 24.2          & 24.3          & 23.2          \\
                         & TRADES$_{\text{CAFE}}$& 80.7          & \textbf{56.6} & \textbf{51.4} & \textbf{48.5} & \textbf{50.4} & \textbf{48.3} & \textbf{46.7} & 91.3          & \textbf{73.9} & \textbf{59.6} & \textbf{54.1} & \textbf{56.7} & \textbf{53.2} & \textbf{51.3} & 54.5          & \textbf{29.6} & \textbf{27.4} & \textbf{26.3} & \textbf{26.5} & \textbf{26.2} & \textbf{25.4} \\
\cmidrule{2-23}
                         & MART                    & \textbf{83.5} & 56.1          & 50.1          & 47.1          & 48.3          & 47.0          & 45.5          & \textbf{93.7} & 74.2          & 58.3          & 51.7          & 53.2          & 50.8          & 47.8          & \textbf{57.1} & 27.4          & 24.2          & 23.2          & 22.9          & 23.2          & 22.2          \\
                         & MART$_{\text{CAFE}}$  & 82.1          & \textbf{57.3} & \textbf{51.9} & \textbf{48.1} & \textbf{50.2} & \textbf{48.0} & \textbf{46.2} & 92.2          & \textbf{74.9} & \textbf{61.0} & \textbf{53.4} & \textbf{57.3} & \textbf{51.8} & \textbf{49.7} & 55.9          & \textbf{28.6} & \textbf{25.9} & \textbf{24.6} & \textbf{24.7} & \textbf{24.5} & \textbf{23.5} \\
\cmidrule{2-23}
                         & AWP                     & 81.2          & 55.3          & 51.6          & 48.0          & 50.5          & 47.8          & 46.9          & 92.2          & 71.1          & 59.8          & 54.3          & 56.8          & 53.6          & 52.0          & 56.2          & 30.5          & 28.5          & 26.2          & 27.6          & 26.2          & 25.5          \\
                         & AWP$_{\text{CAFE}}$   & \textbf{81.5} & \textbf{57.8} & \textbf{54.2} & \textbf{49.4} & \textbf{52.9} & \textbf{49.0} & \textbf{47.8} & \textbf{93.4} & \textbf{74.0} & \textbf{60.9} & \textbf{55.0} & \textbf{57.8} & \textbf{54.8} & \textbf{52.7} & \textbf{56.6} & \textbf{31.4} & \textbf{29.2} & \textbf{27.1} & \textbf{28.4} & \textbf{27.0} & \textbf{26.5} \\
\cmidrule{2-23}
                         & HELP                    & 80.5          & 55.8          & 52.1          & 48.4          & 51.1          & 48.5          & 47.4          & 92.6          & 72.0          & 59.8          & 54.4          & 56.6          & 53.9          & 52.0          & \textbf{56.1} & 31.0          & 28.6          & 26.3          & 27.7          & 26.3          & 25.7          \\
                         & HELP$_{\text{CAFE}}$  & \textbf{80.6} & \textbf{57.8} & \textbf{54.5} & \textbf{49.4} & \textbf{53.1} & \textbf{49.5} & \textbf{48.5} & \textbf{92.9} & \textbf{73.9} & \textbf{61.3} & \textbf{55.3} & \textbf{58.8} & \textbf{54.6} & \textbf{52.8} & 55.4          & \textbf{32.0} & \textbf{29.7} & \textbf{27.4} & \textbf{29.2} & \textbf{27.8} & \textbf{27.3} \\
\midrule

\multirow{10}{*}{\rotatebox[origin=c]{90}{WRN}}{\hskip 1ex}    & ADV                     & 84.3          & 54.5          & 48.7          & 47.8          & 47.0          & 47.9          & 45.6          & \textbf{94.0} & 71.8          & 56.7          & 53.2          & 51.9          & 52.8          & 49.0 & \textbf{60.9} & 29.8 & 25.5 & 25.8 & 24.2 & 26.0 & 23.9           \\ 
                         & ADV$_{\text{CAFE}}$   & \textbf{85.7} & \textbf{58.5} & \textbf{53.3} & \textbf{51.3} & \textbf{51.8} & \textbf{51.5} & \textbf{49.5} & 93.7          & \textbf{75.7} & \textbf{59.1} & \textbf{54.9} & \textbf{54.0} & \textbf{54.1} & \textbf{50.2} & 60.6 & \textbf{31.1}  & \textbf{27.3}  & \textbf{27.2}  & \textbf{25.8}  & \textbf{27.4}  & \textbf{25.4}  \\ 
\cmidrule{2-23}
                         & TRADES                  & \textbf{86.3} & 57.1          & 52.1          & 50.8          & 50.6          & 50.7          & 49.0          & \textbf{93.8} & 74.0          & 58.1          & 53.9          & 53.0          & 53.4          & 49.9 & \textbf{60.8} & 30.5 & 26.4 & 26.7 & 25.0 & 26.8 & 24.6           \\ 
                         & TRADES$_{\text{CAFE}}$& 83.7          & \textbf{58.6} & \textbf{54.5} & \textbf{52.0} & \textbf{53.2} & \textbf{52.0} & \textbf{50.1} & 92.4          & \textbf{75.6} & \textbf{61.0} & \textbf{55.7} & \textbf{58.0} & \textbf{58.0} & \textbf{53.0} & 60.3 & \textbf{31.7}  & \textbf{28.2}  & \textbf{28.3}  & \textbf{27.0}  & \textbf{28.5}  & \textbf{26.5}  \\ 
\cmidrule{2-23}
                         & MART                    & \textbf{86.5} & 58.5          & 52.6          & 50.0          & 50.7          & 49.9          & 48.0          & \textbf{94.2} & 75.0          & 58.0          & 53.1          & 52.8          & 52.8          & 48.9 & \textbf{60.7} & 29.9 & 25.6 & 25.9 & 24.0 & 25.5 & 23.6           \\ 
                         & MART$_{\text{CAFE}}$  & 85.7          & \textbf{59.8} & \textbf{54.6} & \textbf{51.4} & \textbf{52.7} & \textbf{50.9} & \textbf{49.3} & 93.0          & \textbf{76.5} & \textbf{61.9} & \textbf{54.9} & \textbf{57.2} & \textbf{53.8} & \textbf{50.7} & 60.4 & \textbf{31.2}  & \textbf{27.5}  & \textbf{26.8}  & \textbf{25.5}  & \textbf{27.0}  & \textbf{25.1}  \\ 
\cmidrule{2-23}
                         & AWP                     & 83.7          & 58.0          & 54.7          & 51.3          & 53.7          & 51.2          & 50.1          & 93.2          & 73.4          & 60.8          & 55.9          & 57.5          & 55.5          & 53.6 & \textbf{61.9} & 35.5 & 32.8 & 31.0 & 31.6 & 31.1 & 29.6           \\ 
                         & AWP$_{\text{CAFE}}$   & \textbf{84.6} & \textbf{60.6} & \textbf{56.9} & \textbf{52.4} & \textbf{55.5} & \textbf{52.3} & \textbf{51.1} & \textbf{94.2} & \textbf{76.9} & \textbf{62.7} & \textbf{57.5} & \textbf{59.2} & \textbf{57.1} & \textbf{54.6} & 61.4 & \textbf{36.6}  & \textbf{34.2}  & \textbf{32.3}  & \textbf{33.2}  & \textbf{32.5}  & \textbf{30.8}  \\ 
\cmidrule{2-23}
                         & HELP                    & \textbf{83.8} & 58.6          & 54.9          & 51.6          & 53.8          & 51.6          & 50.3          & 93.5          & 73.4          & 60.8          & 56.5          & 57.6          & 56.1          & 54.0 & \textbf{61.8} & 35.9 & 33.0 & 31.3 & 31.8 & 31.3 & 29.8           \\ 
                         & HELP$_{\text{CAFE}}$  & 83.1          & \textbf{60.5} & \textbf{57.1} & \textbf{52.7} & \textbf{56.0} & \textbf{52.6} & \textbf{51.3} & \textbf{94.0} & \textbf{76.6} & \textbf{62.6} & \textbf{57.7} & \textbf{58.8} & \textbf{57.2} & \textbf{55.0} & 61.1 & \textbf{37.0}  & \textbf{34.7}  & \textbf{32.6}  & \textbf{33.8}  & \textbf{32.8}  & \textbf{31.2}  \\ 
\Xhline{3\arrayrulewidth}                       
\end{tabular}
}
\vspace*{-0.2cm}
\caption{Comparison of adversarial robustness and improvement from CAFE on five defense baselines: ADV, TRADES, MART, AWP, HELP, trained with VGG-16, ResNet-18, WideResNet-34-10 for three datasets under six attacks: FGSM, PGD, CW$_{\infty}$, AP, DLR, AA.}
\vspace*{-0.4cm}
\label{tab:cafe}
\end{table*}


For \textit{Relevance}, when taking a look at the estimation procedure of adversarial feature $T$ such that $T=Z+F_{\text{natural}}$, feature variation $Z$ explicitly has a causal influence on $T$. This is because, in our IV setup, the treatment $T$ is directly estimated by instrument $Z$ given natural features $F_{\text{natural}}$. By using all data samples, we empirically compute Pearson correlation coefficient to prove existence of highly related connection between them as described in the last row of~\cref{tab:ivsetup}. Therefore, our IV satisfies \textit{Relevance} condition.

\section{Inoculating CAusal FEatures for Robustness}
Next, we explain how to efficiently implant the causal features into various defense networks for robust networks. To eliminate spurious correlation of networks derived from the adversary, the simplest approach that we can come up with is utilizing the hypothesis model to enhance the robustness. However, there is a realistic obstacle that it works only when we already identify what is natural inputs and their adversarial examples in inference phase. Therefore, it is not feasible approach to directly exploit the hypothesis model to improve the robustness.

To address it, we introduce an inversion of causal features (\ie causal inversion) reflecting those features on input domain. It takes an advantage of well representing causal features within allowable feature bound regarding network parameters of the preceding sub-network $f_{l}$ for the given adversarial examples. In fact, causal features are manipulated on an intermediate layer by the hypothesis model $h$, thus they are not guaranteed to be on possible feature bound. The causal inversion then serves as a key in resolving it without harming causal prediction much, and its formulation can be written with causal perturbation using distance metric of KL divergence $\mathcal{D}_{\text{KL}}$ as:
\begin{equation}
\label{eq:inversion}
    \delta_{\text{causal}}=\argmin\limits_{\left \| \delta \right \|_{\infty} \leq \gamma} \mathcal{D}_{\text{KL}}\left(f_{l+}(F_{\text{AC}})\mid\mid f(X_\delta)\right),
\end{equation}
where $F_{\text{AC}}$ indicates adversarial causal features distilled by hypothesis model $h$, and $\delta_{\text{causal}}$ denotes causal perturbation to represent causal inversion $X_{\text{causal}}$ such that $X_{\text{causal}}=X + \delta_{\text{causal}}$. Note that, so as not to damage the information of natural input during generating the causal inversion $X_{\text{causal}}$, we constraint the perturbation $\delta$ to $l_{\infty}$ within $\gamma$-ball, as known as perturbation budget, to be human-imperceptible one such that $\left \| \delta \right \|_{\infty} \leq \gamma$. Appendix C shows the statistical distance away from confidence score for model prediction of causal features, compared with that of causal inversion, natural input, and adversarial examples. As long as being capable of handling causal features using the causal inversion such that $\hat{F}_{\text{AC}}=f_{l}(X_{\text{causal}})$, we can now develop how to inoculate \textit{CAusal FEatures (CAFE)} to defense networks as a form of empirical risk minimization (ERM) with small population of perturbation $\epsilon$, as follows:
\begin{equation}
\label{eq:cafe}
    \min\limits_{f\in \mathcal{F}} \E_{\mathcal{S}}\left[\max\limits_{\left \| \epsilon \right \|_{\infty} \leq \gamma}\mathcal{L}_{\text{Defense}}+ \mathcal{D}_{\text{KL}}(f_{l+}(\hat{F}_{\text{AC}})\mid\mid f_{l+}(F_{\text{adv}})) \right],
\end{equation}
where $\mathcal{L}_{\text{Defense}}$ specifies a pre-defined loss such as~\cite{madry2018towards, pmlr-v97-zhang19p, Wang2020Improving, wu2020adversarial, rade2022reducing} for achieving a defense network $f$ on network parameter space $\mathcal{F}$, and $\mathcal{S}$ denotes data samples such that $(X,G) \sim\mathcal{S}$. The rest term represents a causal regularizer serving as \textit{causal inoculation} to make adversarial features $F_{\text{adv}}$ assimilate causal features $F_{\text{AC}}$. Specifically, while $\mathcal{L}_{\text{Defense}}$ robustifies network parameters against adversarial examples, the regularizer helps to hold adversarial features not to stretch out from the possible bound of causal features, thereby providing networks to backdoor path-reduced features dissociated from unknown confounders. More details for training algorithm of CAFE are attached in Appendix E.

\section{Experiments}
\subsection{Implementation and Experimental Details}
We conduct exhaustive experiments on three datasets and three networks to verify generalization in various conditions. For datasets, we take low-dimensional datasets: CIFAR-10~\cite{krizhevsky2009learning}, SVHN~\cite{netzer2011reading}, and a high-dimensional dataset: Tiny-ImageNet~\cite{tiny}. To train the three datasets, we adopt standard networks: VGG-16~\cite{vgg}, ResNet-18~\cite{resnet}, and an advanced large network: WideResNet-34-10~\cite{wideresent}.

For attacks, we use perturbation budget $8/255$ for CIFAR-10, SVHN and $4/255$ for Tiny-ImageNet with two standard attacks: FGSM~\cite{43405}, PGD~\cite{madry2018towards}, and four strong attacks: CW$_{\infty}$~\cite{CW}, and AP (Auto-PGD: step size-free), DLR (Auto-DLR: shift and scaling invariant), AA (Auto-Attack: parameter-free) introduced by \cite{croce2020reliable}. PGD, AP, DLR have $30$ steps with random starts where PGD has step sizes $0.0023$ and $0.0011$ respectively, and AP, DLR have momentum coefficient $\rho=0.75$. CW$_{\infty}$ uses gradient clamping for $l_{\infty}$ with CW objective~\cite{CW} on $\kappa=0$ in $100$ iterations. For defenses, we adopt a standard defense baseline: ADV~\cite{madry2018towards} and four strong defense baselines: TRADES~\cite{pmlr-v97-zhang19p}, MART~\cite{Wang2020Improving}, AWP~\cite{wu2020adversarial}, HELP~\cite{rade2022reducing}. We generate adversarial examples using PGD~\cite{madry2018towards} on perturbation budget $8/255$ where we set $10$ steps and $0.0072$ step size in training. Especially, adversarially training for Tiny-ImageNet is a computational burden, so we employ fast adversarial training~\cite{Wong2020Fast} with FGSM on the budget $4/255$ and its $1.25$ times step size. For all training, we use SGD~\cite{robbins1951stochastic} with $0.9$ momentum and learning rate of $0.1$ scheduled by Cyclic~\cite{smith2017cyclical} in $120$ epochs~\cite{pmlrv119rice20a, Wong2020Fast}.

\subsection{Comparing Adversarial Robustness}
We align the above five defense baselines with our experiment setup to fairly validate adversarial robustness. From~\cref{eq:inversion}, we first acquire causal inversion to straightly deal with causal features. Subsequently, we employ the causal inversion to carry out causal inoculation to all networks by adding the causal regularizer to the pre-defined loss of the defense baselines from scratch, as described in~\cref{eq:cafe}. \cref{tab:cafe} demonstrates CAFE boosts the five defense baselines and outperforms them even on the large network and large dataset, so that we verify injecting causal features works well in all networks. Appendix F shows ablation studies for CAFE without causal inversion to identify where the effectiveness comes from.

\begin{figure}[t!]
\centering
\includegraphics[width=0.99\linewidth]{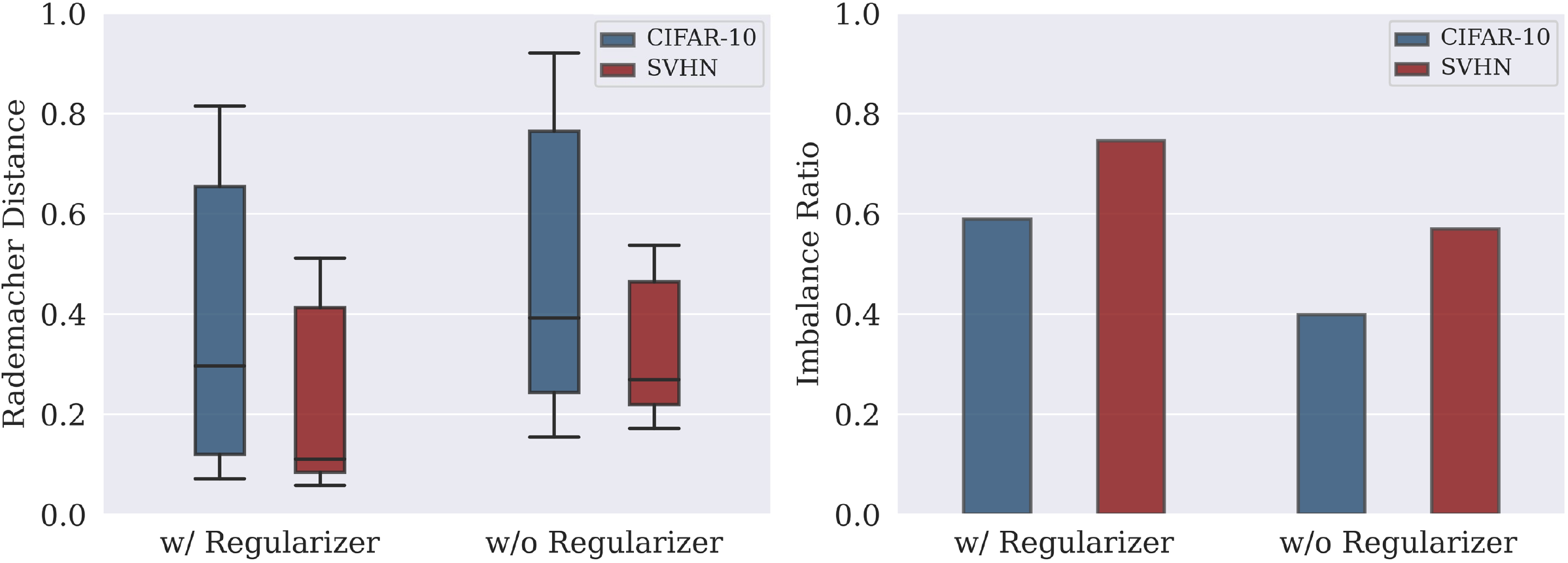}
\vspace*{-0.3cm}
\begin{flushleft}
    \hspace{0.5cm}{(a) Rademacher Distance\hspace{1.0cm}(b) Imbalance Ratio}
\end{flushleft}
\vspace*{-0.4cm}
\caption{Displaying box distribution statistics of Rademacher Distance and Imbalance ratio for prediction results, compared with w/ Regularizer and w/o Regularizer on two datasets for VGG-16.}
\label{fig:rad}
\vspace{-0.6cm}
\end{figure}

\subsection{Ablation Studies on Rich Test Function}
To validate that the regularizer truely works in practice, we measure Rademacher Distance and display its box distribution as illustrated in~\cref{fig:rad} (a). Here, we can apparently observe the existence of the regularization efficiency through narrowed generalization gap. Concretely, both median and average of Rademacher Distance for the regularized test function are smaller than the non-regularized one. Next, in order to investigate how rich test function helps causal inference, we examine imbalance ratio of prediction results for the hypothesis model, which is calculated as \# of minimum predicted classes divided by \# of maximum predicted classes. If the counterfactual space deviates from possible feature bound much, the attainable space that hypothesis model can reach is only restricted areas. Hence, the hypothesis model may predict biased prediction results for the target objects. As our expectation, we can observe the ratio with the regularizer is largely improved than non-regularizer for both datasets as in~\cref{fig:rad} (b). Consequently, we can summarize that rich test function acquired from the localized Rademacher regularizer serves as a key in improving the generalized capacity of causal inference.

\section{Conclusion}
In this paper, we build AMR-GMM to develop adversarial IV regression that effectively demystifies causal features on adversarial examples in order to uncover inexplicable adversarial origin through a causal perspective. By exhaustive analyses, we delve into causal relation of adversarial prediction using hypothesis model and test function, where we identify their semantic information in a human-recognizable way. Further, we introduce causal inversion to handle causal features on possible feature bound of network and propose causal inoculation to implant \textit{CAusal FEatures (CAFE)} into defenses for improving adversarial robustness.


\newpage
{\small
\bibliographystyle{ieee_fullname}
\bibliography{main}
}

\end{document}